\documentclass[10pt,twocolumn,letterpaper]{article}

\usepackage[pagenumbers]{iccv} 
\usepackage{microtype}
\usepackage{subcaption}
\usepackage{booktabs} 
\usepackage{makecell}

\usepackage[OT1]{fontenc} 
\usepackage{times}
\usepackage{amsmath}
\usepackage{amssymb}
\usepackage{mathtools}
\usepackage{amsthm}
\usepackage{url}
\usepackage{epsfig}
\usepackage{graphicx}
\usepackage{xcolor}
\usepackage{multirow}
\usepackage{comment}
\usepackage{tabularx}
\usepackage{enumitem}
\usepackage{caption}
\usepackage{diagbox}

\newcommand{\imh}{\mathrm{H}}
\newcommand{\imw}{\mathrm{W}}

\definecolor{cvprblue}{rgb}{0.21,0.49,0.74}
\usepackage[pagebackref,breaklinks,colorlinks,allcolors=cvprblue]{hyperref}

\DeclareMathOperator*{\argmin}{arg\,min}

\begin{document}

\title{Divided Attention: Unsupervised Multi-object Discovery by Motion with Contextually Separated Slots}


\author{Dong Lao\thanks{Correspondence: lao@cs.ucla.edu and yanchaoy@hku.hk.}\\UCLA
\and
Zhengyang Hu\\HKU
\and
Francesco Locatello\\ISTA
\and
Yanchao Yang\footnotemark[1]\\HKU
\and
Stefano Soatto\\UCLA
}

\maketitle

\begin{abstract}
We investigate the emergence of objects in visual perception in the absence of any semantic annotation. The resulting model has received no supervision, does not use any pre-trained features, and yet it can segment the domain of an image into \emph{multiple} independently moving regions. The resulting motion segmentation method can handle an unknown and varying number of objects in real-time. The core multi-modal conditional encoder-decoder architecture has one modality (optical flow) feed the encoder to produce a collection of latent codes (slots), and the other modality (color image) conditions the decoder to generate the first modality (flow) from the slots. The training criterion is designed to foster `information separation' among the slots, while the architecture explicitly allocates activations to individual slots, leading to a method we call Divided Attention (DivA). At test time, DivA handles a different number of objects and different image resolution than seen at training, and is invariant to permutations of the slots. DivA achieves state-of-the-art performance while tripling run-time speed of comparable methods up to 104FPS, and reduces the performance gap from supervised methods to 12\% or less. Objects bootstrapped by DivA can then be used to prime static classifiers via contrastive learning. On fewer than 5,000 video clips, training DINO on DivA’s object proposals narrows the performance gap to ImageNet-based training by up to 30.2\% compared to training directly on the video frames.
\end{abstract}

\section{Introduction}
\label{sec:introduction}

\def\figd{figures/fbms}
\def\fWidD{0.12\textwidth}
\begin{figure*}[t]
\renewcommand{\arraystretch}{0.2} 
\centering
{\footnotesize
\begin{subfigure}{\textwidth}
\hspace*{-0.1in}
\begin{tabular}[0mm]{c@{\hskip 0.01in}c@{\hskip 0.01in}c@{\hskip 0.1in}c@{\hskip 0.01in}c@{\hskip 0.01in}c@{\hskip 0.01in}c@{\hskip 0.1in}c}
Image&Flow&Recon. Flow&Slot 1&Slot 2&Slot 3&Slot 4&Segmentation \\
\includegraphics[width=\fWidD]{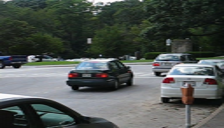}&
\includegraphics[width=\fWidD]{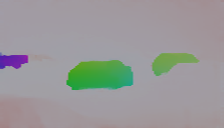}&\includegraphics[width=\fWidD]{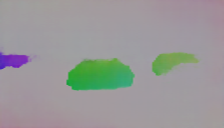}& 
\includegraphics[width=\fWidD]{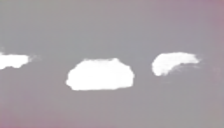} &
\includegraphics[width=\fWidD]{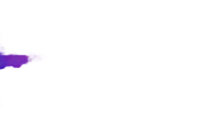} &
\includegraphics[width=\fWidD]{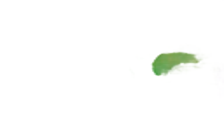} &
\includegraphics[width=\fWidD]{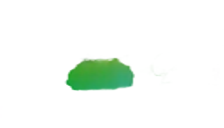}& 
\includegraphics[width=\fWidD]{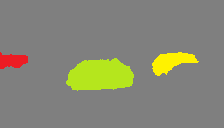}\\
\includegraphics[width=\fWidD]{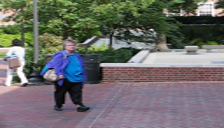}&
\includegraphics[width=\fWidD]{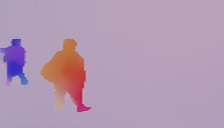}&\includegraphics[width=\fWidD]{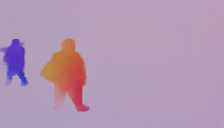}& 
\includegraphics[width=\fWidD]{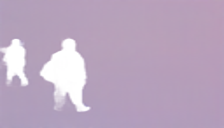} &
\includegraphics[width=\fWidD]{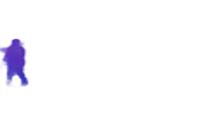} &
\includegraphics[width=\fWidD]{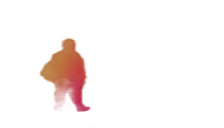} &
\includegraphics[width=\fWidD]{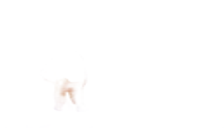}& 

\includegraphics[width=\fWidD]{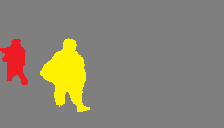}\\
\includegraphics[width=\fWidD]{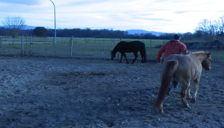}&
\includegraphics[width=\fWidD]{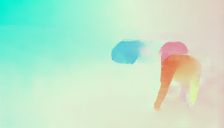}&\includegraphics[width=\fWidD]{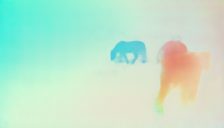}& 
\includegraphics[width=\fWidD]{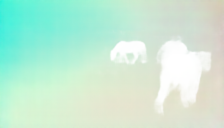} &
\includegraphics[width=\fWidD]{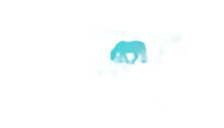} &
\includegraphics[width=\fWidD]{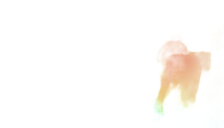} &
\includegraphics[width=\fWidD]{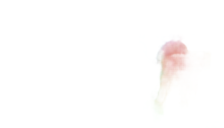}& 

\includegraphics[width=\fWidD]{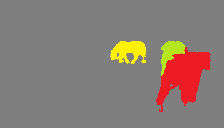}\\
\includegraphics[width=\fWidD]{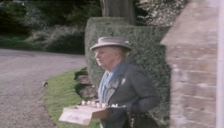}&
\includegraphics[width=\fWidD]{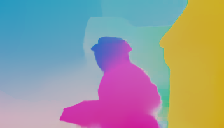}&\includegraphics[width=\fWidD]{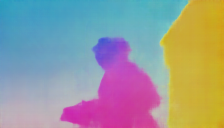}& 
\includegraphics[width=\fWidD]{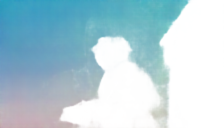} &
\includegraphics[width=\fWidD]{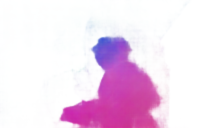} &
\includegraphics[width=\fWidD]{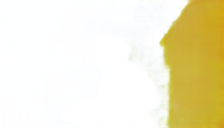} &
\includegraphics[width=\fWidD]{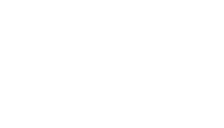}& 
\includegraphics[width=\fWidD]{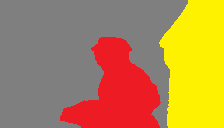}\\
\end{tabular}
   \caption{\small Multi-region segmentation by motion on real-world data. Dataset: FBMS-59.}
\label{fig:fbms}
\end{subfigure}
\begin{subfigure}{\textwidth}
\vspace*{-0.15in}
\hspace*{0.3in}
\includegraphics[width=0.92\textwidth]{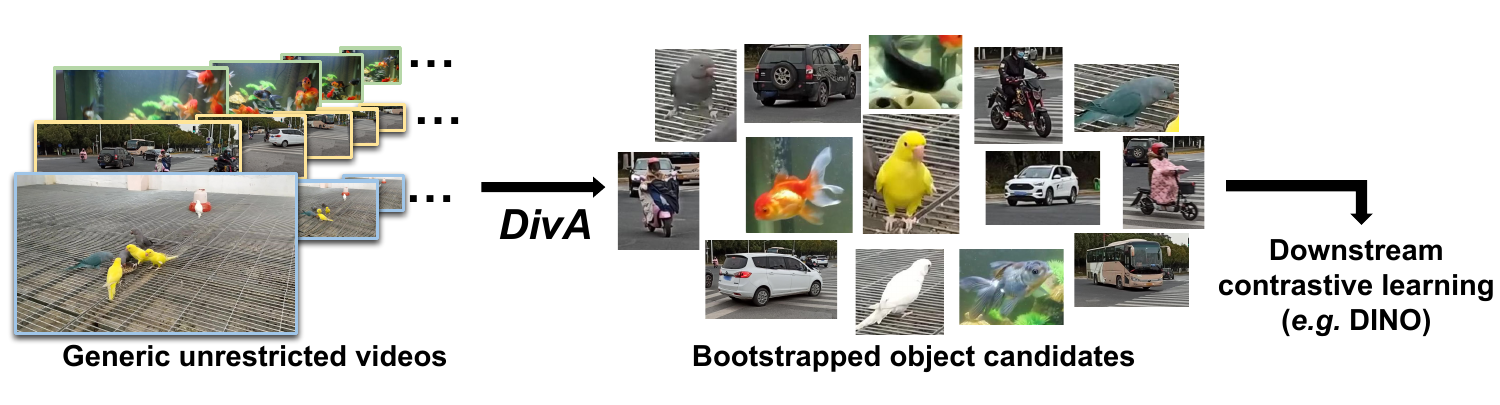}
\vspace*{-0.12in}
   \caption{\small Fully unsupervised object bootstrapping and representation learning pipeline. Dataset: MOSE.}
\label{fig:dino}
\end{subfigure}
\vspace*{-0.27in}
\caption{\sl\small {\bf Overview.} {\bf (a)} Given videos, DivA segments the image domain into independently moving regions without pre-trained features or human annotation. The slots are randomly initialized, so as to be permutable. In the final row, DivA segments a stationary wall due to parallax-induced motion, demonstrating how motion may enable object bootstrapping in embodied systems, even if the object is stationary. {\bf (b)} DivA receives an unordered set of variable-length videos and outputs a set of discovered objects, without any human prior. 
The set of discovered objects can serve as initial candidates for training a foundation model which can subsequently be applied to static images.}}
\vspace*{-0.1in}


\end{figure*}

We investigate how object perception may develop \emph{ab-ovo} in vision systems, priming  semantic understanding of the surrounding environment, by 
exploring object representations in the limit of minimal to no supervision. 

Recent advancements in large-scale vision foundation models have mirrored the process used in language models, effectively reversing evolution by leveraging vast amounts of manual semantic supervision \cite{kirillov2023segment,ravi2024sam} including captions \cite{radford2021learning} or purposeful framing of a `dominant object' in each image \cite{caron2021emerging, oquab2023dinov2}). We do not follow this trend, and instead ask how object perception may arise from a simple constitutive bias (loss function), leveraging  the structure of the surrounding environment  \cite{lao2024on} and so-called `ecological statistics' \cite{gibson1978ecological} without any explicit supervision beyond the design of the learning criterion.
In biology, visual skills emerge without symbolic supervision,  while the ability to detect objects by motion is crucial to survival and shared with our reptilian ancestors \cite{chalupka2016generalized}. We are curious to explore such a bootstrapping process in an artificial setting, 
by leaving assigning symbols to objects as a post-hoc ``alignment'' of naming.

Aside from scientific curiosity, in practice there are still plenty of domains where data is scarce and different enough from web-scraped data that large-scale foundation models,  designed to `always work,' don't. These include embodied systems with non-conventional sensors, biological imagery, domains where the data is scarce due to privacy constraints, planetary exploration, etc.. In these scenarios, an unsupervised object discovery approach free of manual annotation and image capture biases would be desirable. 

We have thus far refrained from formally defining  `objects.' We adopt Gibson's definition of ``detached objects'' \cite{gibson1978ecological} and variants \cite{ayvaci2011detachable}, and aim to segment the visual field into {\em regions of an image whose corresponding motion is unpredictable from their surroundings.} This definition may initially seem counterintuitive: A person sitting in a car or a cup resting on a table is treated as (multiple connected components of) a single entity since they move coherently. However, as time goes by, once the person leaves the car or the cup is moved, they would be revealed as distinct objects.

Such a ``separation'' criterion offers us a principle for object segmentation - what moves together belongs together - that does not rely on an artificial symbolic hierarchy for partitioning, and can be implemented as a variational principle known as Contextual Information Separation (CIS) \cite{yang2019unsupervised}, that is the division of the image domain into regions whose motions are \emph{mutually uninformative}. However, the CIS criterion has thus far only been applied to binary segmentation, relying on the data to feature a prominent `foreground' object. Removing this selection bias and generalizing to \emph{multiple} and an \emph{unknown} numbers of partitions is non-trivial, thus we also impose an architectural bias, using Slot Attention Networks (SAN) \cite{locatello2020object} that explicitly allocate different activations to different `slots' in the input data. However, naively applying SANs to unrestricted real-world data yields slot collapse, preventing it from working on highly complex scenes. We overcome this limitation by performing {\em cross-modal generation}:
the model reconstructs motion (optical flow, which has less nuisance variability than the image), but uses the image to {\em guide} flow reconstruction. Ideally, each slot would be interpretable as an independent region, but since SANs are trained by a reconstruction loss without an explicit bias to assign independent regions to different slots, multiple slots can correspond to the same region (slot collapse). We therefore use an adversarial scheme to foster  {\em `divided attention'}.  The resulting method, DivA (Fig.~\ref{fig:schematic}), consists of a multi-modal architecture where one modality (flow) goes into the encoder that {\em feeds} the slots, and the other (image) goes into the decoder to {\em guide} the output, trained adversarially with a loss (\ref{eq:loss}) that captures both cross-modal generation and information separation. To our knowledge, DivA is the first motion segmentation and object discovery method that does not use any (implicit or explicit) supervision or pre-processing, besides the choice of architecture and training loss, yet it can: 

\setlist{nosep}
\begin{itemize}[leftmargin=*]
    \item operates in real-time and requires only single images and corresponding flow as input;
    \item does not rely on object-centric pre-training data or require any pre-trained image features from external datasets.
    \item seamlessly handles a variable number of objects that can be changed at inference time without needing to re-train, and is invariant to permutations of object ordering;
\end{itemize}
While these properties are consistent with bootstrapping in evolutionary development, an added benefit is interpretable results: the adversarial loss fosters the coalescence of pixels and the separation of objects in different slots. 

DivA demonstrates efficacy in multi-object motion segmentation on both synthetic (Movi-Solid) and real-world (FBMS-59) videos. On commonly-used binary benchmarks, DivA outperforms baselines on DAVIS and SegTrack by 5\% and 7\% respectively, and reduces the gap from batch-processing and supervised methods to 9.5\% and 12\% respectively. Compared to these paragon methods, DivA has low inference latency, as it does not need to process a video batch to produce an outcome, and low computational cost enables real-time operation: An embodiment of DiVA that matches the current state-of-the-art improves speed by 200\% (21 FPS to 64 FPS), and our fastest embodiment reaches 104 FPS.

\subsection{Related Work}
\label{sec:related}

\textbf{Unsupervised motion segmentation. }
Early methods segment dense optical flow \cite{horn1981determining} by optimizing a designed objective through the evolution of a partial differential equation (PDE) \cite{memin2002hierarchical
},  decomposing it into layers  \cite{wang1994representing, weiss1996unified, sun2010layered, lao2018extending}, or grouping  sparse trajectories \cite{ochs2013segmentation,keuper2015motion,keuper2016multi}. The number of clusters/layers/objects was determined by regularization parameters or explicitly enforced, and these methods require solving complex optimization problems at inference time. 

Deep neural networks often segment an image into pre-defined classes corresponding to output channels \cite{noh2015learning,zhao2017pyramid,chen2017deeplab}. Many ``binary'' methods \cite{Tokmakov17, fusionseg, song2018pyramid, liu2021emergence, ding2022motion, lian2023bootstrapping, singh2023locate} are even more restrictive, considering two classes: ``foreground'' and ``background'' 
\cite{hu2018unsupervised, Fan_2019_CVPR, Li_2019_ICCV, Akhter_2020_WACV}. Some methods revisit layered models by supervising a DNN using synthetic data  \cite{xie2022segmenting} or use sprites \cite{ye2022sprites}. A video batch must be processed before inference on a given frame, and the model requires manual input of the number of regions, rendering these approaches impractical in online and real-time scenarios. \cite{meunier2022driven} segments optical flow by fitting pre-defined motion patterns using Expectation-Maximization (EM). Changing the number of regions requires architectural modification and thus re-training. {\color{black} DivA is more flexible by only specifying an upper limit for the number of segments (slots) which can be conveniently adjusted without re-training the model.}

Some unsupervised object discovery methods \cite{wang2022tokencut, ponimatkin2023simple, bielski2022move,seitzer2022bridging, zadaianchuk2023object, wang2023cut} use pre-trained image features from contrastive learning (e.g. DINO \cite{caron2021emerging}), while others \cite{bao2022discovering,bao2023object} rely on motion segmentation \cite{dave2019towards} that leverages supervised features from MS-COCO \cite{lin2014microsoft}. There are also methods combining an unsupervised pipeline with supervised features \cite{yang2021dystab, haller2021iterative}. 
While in practice one ought to use pre-trained features whenever available, for academic purposes we explore the minimal end of the supervision scale, where no purposeful viewpoint selection, ``primary object,'' or human labels are used. %


\textbf{Contextual Information Separation (CIS)}
\cite{yang2019unsupervised} frames unsupervised object discovery as an information separation task by segmenting the optical flow field into two regions that are muturally uninformative. Although this formulation removes the notion of foreground and background, it is still limited to binary partitioning. No work has yet generalized CIS to multiple objects. Optimizing a multi-region version of CIS would result in a combinatorial explosion of complexity, while naive alternatives such as sequentially discovering one object at a time did not yield satisfactory results, either. 

\textbf{Slot Attention Networks (SANs)} \cite{locatello2020object} infer a set of latent variables each ideally representing an object in the image. Earlier ``object-centric learning'' methods \cite{eslami2016attend,greff2017neural,kosiorek2018sequential,greff2019multi} aim to discover generative factors that correspond to parts or objects in the scene, but require solving an iterative optimization at test time; SAN processes the data in a feed-forward pass by leveraging the attention mechanism \cite{vaswani2017attention} that allocates latent variables to a collection of permutation-invariant slots. SAN is trained to minimize the reconstruction loss, with no explicit mechanism enforcing the separation of objects. When objects have distinct features, the low-capacity bottleneck is sufficient to separate objects into slots. However, in realistic images, multiple objects tend to collapse to the same slots (Fig.~\ref{fig:movi}), so a more explicit bias than that implicit in the architecture is needed. To that end,  \cite{kipf2021conditional} resorts to external cues such as bounding boxes,  while \cite{elsayedsavi++} employs Lidar. Instead, DivA modifies the SAN architecture by using a cross-modal conditional decoder inspired by \cite{yang2018conditional}. By combining
information separation, architectural separation, and conditional generation, DivA fosters better partition of the slots into objects. DivA is most closely related to MoSeg \cite{yang2021self}, which adopts SAN for iterative binding to the flow. DivA has two key advantages: no dependency on learned slots initializations (Fig.~\ref{fig:inverted}), and cross-modal reconstruction through a conditional decoder while MoSeg only uses flow.

\section{Method}
DivA ingests a color (RGB) image $I \in {\mathbb R}^{\imh\times\imw\times 3}$ and its associated optical flow $u \in {\mathbb R}^{\imh\times\imw\times 2}$ defined on the same lattice (image domain). DivA outputs a collection of $n$ masks $m_i, \ i = 1, \dots, n$ and the reconstructed flow $\hat u$.

\begin{figure*}[t]
\centering
\hspace*{1cm}
\includegraphics[width=0.87\textwidth]{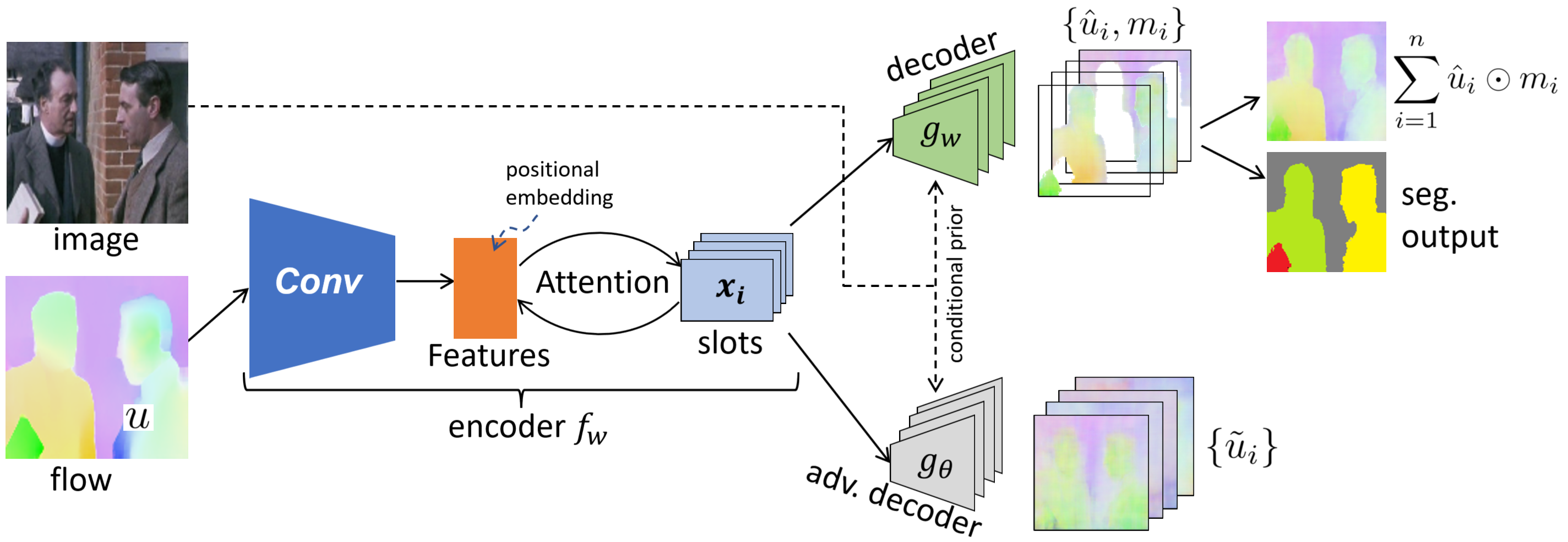}
\vspace*{-0.1in}
    \caption{\sl\small {\bf Divided Attention architecture.}
        Unlike traditional autoencoders that process the input (here the flow $u$) through an encoder $f_w$, and then reconstruct it through a decoder $g$, DivA uses a cross-modal conditional decoder ($g_w$ in green) that takes a second modality as input (here the image $I$) as a conditional prior to guide the decoder to reconstruct the first from the ``slots'' $x_i$ (light blue). To ensure that individual slots encode separate objects, we design an adversarial decoder ($g_\theta$ in grey) that tries to reconstruct the entire flow with each slot. Training is done by optimizing a min-max criterion whereby the model tries to reconstruct the input flow within each object mask, while fooling the adversarial decoder outside. This design enforces information separation between slots, leading to Divided Attention.}
    \label{fig:schematic}\vspace*{-0.1in}
\end{figure*}

\subsection{Preliminaries: Slot Attention Autoencoder}
\label{sec:slot}

The DivA architecture comprises a collection of latent codes $X_n = \{x_1, \dots, x_n\}$, with each $x_i \in {\mathbb R}^{1\times K}$ 
representing a ``slot.'' The encoder $f_w(u)= X_n$ is the same as a Slot Attention Network (SAN), described in detail in \cite{locatello2020object} and summarized here. SANs are trained as autoencoders: An image $I$ is passed through a CNN backbone with an appended positional embedding, but instead of having a single latent vector, SAN uses a collection of them $\{x_i|i = 1,\cdots, n\}$, called \emph{slots}, in the bottleneck, where $n$ may change anytime during training or testing. Slots are initially sampled independently from a Gaussian with a learned mean and standard deviation. This affords changing the number of slots without re-training, and yields invariance to permutation the slots, which are updated iteratively using dot-product attention normalized over slots fostering competition among slots. The result is passed through a Gated Recurrent Unit (GRU) with a multi-layer perceptron (MLP) to yield the update residual for slots. All parameters are shared among slots to preserve permutation symmetry. Each slot is then decoded independently with a spatial broadcast decoder~\cite{watters2019spatial} $g$ producing slot reconstructions $\hat{I}_i$'s and masks $m_i$'s. The final image reconstruction is $\hat{I} = \sum_{i=1}^n \hat{I}_i\odot m_i$ where $\odot$ denotes element-wise multiplication. SAN is trained by minimizing a reconstruction loss (typically MSE) between $I$ and $\hat{I}$. 

\subsection{Cross-modal Conditional Slot Decoder}
\label{sec:conditional}

Experimental evidence shows that slots can learn representations of independent simple objects on synthetic images. However, the naive use of SAN to jointly auto-encode real-world images and flow leads to poor reconstruction.  Since the combined input is complex and the slots are low-dimensional, slots tend to either approximate the entire input or segment the image in a grid pattern that ignores the objects. Both lead to poor separation of objects.

For these reasons, we choose {\em not} to use the image as input to be reconstructed, but as {\em context} to condition the reconstruction of a simpler modality -- the one least affected by complex nuisance variability -- flow in our case. The conditional decoder $g_w$ maps each  latent code $x_i$ {\em and} image $I$ onto a reconstructed flow and a mask $\{\hat u_i, m_i\} = g_u(x_i, I)$. The detailed architecture can be found in the Supp. Mat.

This decoder performs cross-modal transfer since the image serves as a prior for generating the flow. This is akin to a Conditional Prior Network \cite{yang2018conditional}, but instead of reconstructing the entire flow, we reconstruct individual flow components corresponding to objects in the scene, indicated by the masks. The flow is reconstructed by $\hat u = \sum_{i=1}^n \hat u_i \odot m_i$.
Next, we further use an {\em adversarial conditional decoder} to generate $\tilde{u} = g_\theta (x_i, I)$, that attempts to reconstruct the entire flow $\tilde u$ {\em from each individual slot} $x_i$, which in turn encourages the separation between slots.

\subsection{Adversarial Learning}
\label{sec:cis}

We now describe the separation criterion, which we derive from CIS \cite{yang2019unsupervised}, to divide slots into objects. Each image $I$ and flow $u$ in the training set contribute a term in the loss:
 \begin{equation}
        \ell(u, I) = \big\| u - \sum_{i=1}^n \hat u_i \odot m_i \| -  \frac{\lambda}{n}\sum_{i=1}^n (1-m_i ) \odot \| u - \tilde u_i \|
    \end{equation}

The first term penalizes the reconstruction error by the cross-modal autoencoder combining all the slots, i.e., we want the reconstructed flow to be as good as possible. The second term combines the reconstruction error of the adversarial decoder using a single slot at a time, and maximizes its error outside the object mask. Note that, in the second term, the adversarial decoder $g_\theta$ tries to approximate the entire flow $u$ with a single slot $\tilde u_i = g_\theta(x_i, I)$, which is equivalent to maximizing the contextual information separation between different slots.
We use the mean-square reconstruction error (MSE) $d(u,v) = \| u-v \|_2$ for simplicity, but one can replace it with any other distance or discrepancy measure such as empirical cross-entropy, without altering the core method.
The overall loss over training samples in a dataset $D$, is then minimized with respect to the parameters $w$ of the encoder $f_w$ and decoder $g_w$, and  maximized with respect to the parameters $\theta$ of the adversarial decoder $g_\theta$:
\begin{equation}\label{eq:loss}
    \min_{w}\max_{\theta} L(w,\theta ) = \sum_{(u_j, I_j)\in D} \ell(u_j, I_j).
\end{equation}

DivA is minimizing mutual information {\em between different slots}, where the data is encoded, rather than directly {\em between different regions}, which eliminates degenerate slots thanks to the reconstruction objective. The resulting design is a natural extension of CIS to non-binary partitions, and also leads to increased versatility: DivA can be trained with a certain number of slots, on images of a certain size, and used at inference time with a different number of slots, on images of different size. Note that the loss does not need explicit regularization, although one can add it if so desired.

\section{Experiments}

\def\figd{figures/adv}
\begin{figure}[t]
\centering
\begin{subfigure}{0.25\textwidth}
\footnotesize
 \begin{tabular}{l|cc}
Method&bIoU&SPC\\
\hline
Vanilla SAN&51.18&120\\
SAN + Cond.&82.93&133\\
DivA ($\lambda=0.01$)&\bf{84.49}&161\\
DivA ($\lambda=0.03$)&82.33&\bf{182}\\
DivA ($\lambda=0.05$)&79.56&174
\end{tabular}
\caption{}
\label{tab:adv}
\end{subfigure}
\begin{subfigure}{0.21\textwidth}
\hspace*{1mm}
\vspace*{-1mm}
\includegraphics[width=0.92\textwidth]{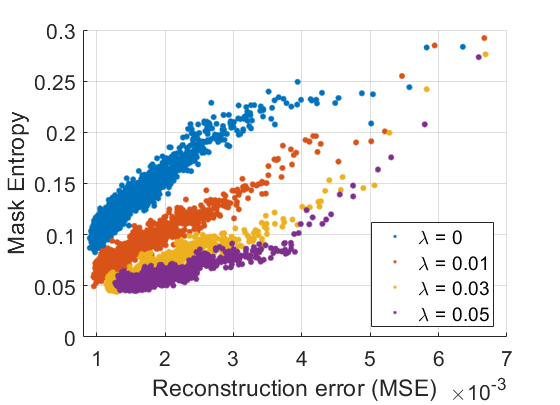}
\caption{}
\label{fig:adv}
\end{subfigure}
\vspace*{-3mm}
\caption{\sl\small {\bf Results on diagnostic data.} (a) Conditional decoder improves bootstrapping IoU (bIoU), indicating better awareness of object shape; adversarial training improves successful partition count (SPC), indicating better information separation in slots. (b) At each reconstruction error level, the higher $\lambda$ we apply, the smaller entropy we get in the segmentation output, rendering adversarial loss an implicit entropy regularizer.}
\vspace*{-0.1in}
\end{figure}
\textbf{The encoder} $f$ consists of 4 convolutional layers with kernel size equal to $5\times5$, consistent with the original implementation of \cite{locatello2020object}. With padding, the feature map keeps a same spatial resolution as the network input. Since motion is less complex than images, we choose $K=48$ instead of 64 used by the original SAN, leading to a narrower bottleneck. A learned positional embedding is then added to the feature map. We keep the iterative update of slots the same as the original SAN, and fix the number of slot iterations to be 3.

\textbf{The conditional decoder}
 $g_w$ (see Supp. Mat.) consists of two parts: an image encoder, and a flow decoder. We use 5 convolutional layers in the image encoder. Same as $f$, with padding, the size of the feature maps remains the same as $I$. We limit the capacity of this encoder by setting the output channel dimension of each layer to 24 to avoid overfitting the flow to the image. The flow decoder takes one $1\times48$ slot vector $x_i$ as input. It first broadcasts $x_i$ spatially to $h\times w\times48$, and adds it to a learned positional embedding, different from the one in $f$. The broadcasted slot then passes through 6 convolutional layers. The feature map at each layer is concatenated with the image feature at the corresponding level. The last convolutional layer outputs a $h\times w\times3$ field, where the first 2 channels reconstruct the optical flow, and the 3rd channel outputs a mask. The adversarial decoder shares the same architecture, except for the last layer, which outputs 2 channels aiming to reconstruct the flow on the entire image domain.

\textbf{Optimization.} We aim to keep the training pipeline simple and all models are trained on a single Nvidia 1080Ti GPU with PyTorch. We implement implicit differentiation during training \cite{chang2022object} for stability, and apply alternating optimization to $w$ and $\theta$ by Eq.~\eqref{eq:loss}. In each iteration, we first fix $g_{\theta}$ and update $w$, then use \textit{torch.detach()} to stop gradient computation on $x_i$ before updating $\theta$. This makes sure that only $\theta$ is updated when training the adversarial decoder. We train with batch size 32 and apply the ADAM optimizer with an initial learning rate $8e^{-4}$ for both $w$ and $\theta$. We notice the model is not sensitive to the choice of learning rate.

\subsection{Results} \label{sec:results}

\textbf{Diagnostic data and ablation.} Due to the limited availability of benchmark datasets for multiple moving objects, we reproduce the ideal conditions of the experiments reported in \cite{yang2019unsupervised} to perform controlled ablation studies. We generate flow-image pairs that contain $n = 2,3,4$ regions with statistically independent motion patterns. We adopt object masks from DAVIS and paste them onto complex background images so that the network cannot overfit to image appearance for reconstruction and segmentation. During training and testing, we fix the number of slots to be 4.

We evaluate on 300 validation samples and measure the performance by \emph{bootstrapping IoU (bIoU)} and \emph{successful partition counts (SPC)}. bIoU is computed by matching each ground truth object mask to the segment with the highest IoU, and averaging this IoU across all objects in the dataset. It measures how successfully each object is bootstrapped. As bIoU does not penalize falsely bootstrapped blobs, in addition, SPC counts the number of cases where the number of objects in the segmentation output matches the ground truth number of objects. As multiple slots may map to the same object, a higher SPC is achieved when information is separated among slots, reducing such confusion.

We apply SAN to \emph{reconstruct the flow} as the baseline 
, and also combining SAN with the conditional decoder. Fig.~\ref{tab:adv} summarizes the results. Our conditional decoder allows exploiting photometric information, in addition to motion, improving bIoU; adversarial learning fosters better separation among slots, reducing slot confusion and improving SPC. In Fig.~\ref{fig:adv} we also display the scatter between reconstruction error and segmentation $\text{Mask Entropy} = \sum_i^4 m_i \log(m_i)$ during training. A smaller entropy corresponds to a more certain mask output, indicating that the decoder relies more on single slots to reconstruct optical flow for each particular object. At each level of reconstruction error, the larger $\lambda$, the smaller the mask entropy. This phenomenon validates the effect of our adversarial training. Note that entropy regularization is also applied to existing unsupervised segmentation methods ({\em e.g.} MoSeg), and our adversarial loss provides a potential alternative to entropy regularization.

\begin{table}[t]
\footnotesize
  \centering
  \hspace*{-0.1in}
  \begin{tabular}{l|ccc|ccc}
  &\multicolumn{3}{c|}{$\delta t = 1$}&\multicolumn{3}{c}{$\delta t = 2$}\\
\hline    
    n&SAN&w/ cond.&DivA&SAN&w/ cond.&DivA\\
      \hline
    3&17.76&39.81&40.87&17.66&40.90&41.27 \\
    4&19.27&40.96&\textbf{42.54}&19.57&39.81&\textbf{41.85} \\
    5&20.17&38.73&40.25&20.15&38.47&40.48 \\
    6&19.69&38.03&38.90&17.90&38.59&40.26 
    \end{tabular}
  \vspace*{-0.1in}
  \caption{\sl\small{\bf Bootstrapping accuracy on FBMS-59. } We evaluate the performance on FBMS-59 by bIoU, varying the number of slots (without re-training). We report results on the baseline SAN \cite{locatello2020object}, SAN with our conditional decoder, and the full DivA model.}
    \label{tab:fbms}
\vspace*{-0.1in}
\end{table}

\def\figd{figures/movi}
\def\fWidD{0.09\textwidth}
\begin{figure}[t]

\renewcommand{\arraystretch}{0.2} 
\centering
\hspace*{-3mm}
{\scriptsize
\begin{tabular}[0mm]{c@{\hskip 0.01in}c@{\hskip 0.01in}c@{\hskip 0.01in}c@{\hskip 0.01in}c@{\hskip 0.01in}c}
\smash{\rotatebox{90}{\,\,\quad Image}}&
\includegraphics[width=\fWidD]{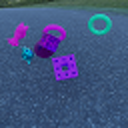}&
\includegraphics[width=\fWidD]{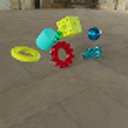}&
\includegraphics[width=\fWidD]{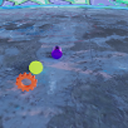}&
\includegraphics[width=\fWidD]{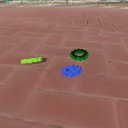}&
\includegraphics[width=\fWidD]{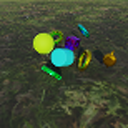}\\
\smash{\rotatebox{90}{\quad\,\,Flow}}&
\includegraphics[width=\fWidD]{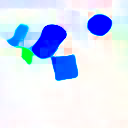}&
\includegraphics[width=\fWidD]{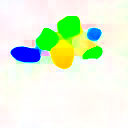}&
\includegraphics[width=\fWidD]{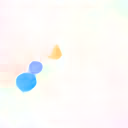}&
\includegraphics[width=\fWidD]{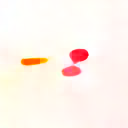}&
\includegraphics[width=\fWidD]{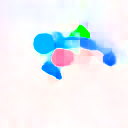}\\
\smash{\rotatebox{90}{\quad\,\,DivA}}&
\includegraphics[width=\fWidD]{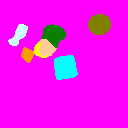}&
\includegraphics[width=\fWidD]{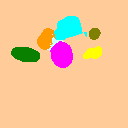}&
\includegraphics[width=\fWidD]{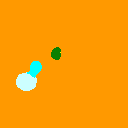}&
\includegraphics[width=\fWidD]{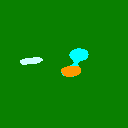}&
\includegraphics[width=\fWidD]{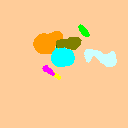}\\
\smash{\rotatebox{90}{SAN+Cond.}}&
\includegraphics[width=\fWidD]{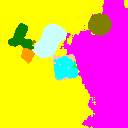}&
\includegraphics[width=\fWidD]{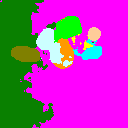}&
\includegraphics[width=\fWidD]{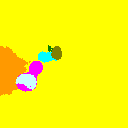}&
\includegraphics[width=\fWidD]{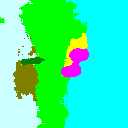}&
\includegraphics[width=\fWidD]{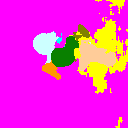}\\
\smash{\rotatebox{90}{\quad\,\,SAN}}&
\includegraphics[width=\fWidD]{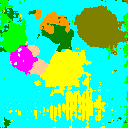}&
\includegraphics[width=\fWidD]{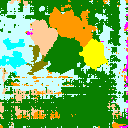}&
\includegraphics[width=\fWidD]{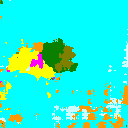}&
\includegraphics[width=\fWidD]{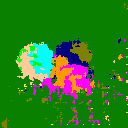}&
\includegraphics[width=\fWidD]{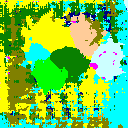}\\
\smash{\rotatebox{90}{\,\,SAN-img}}&
\includegraphics[width=\fWidD]{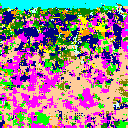}&
\includegraphics[width=\fWidD]{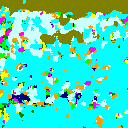}&
\includegraphics[width=\fWidD]{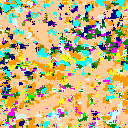}&
\includegraphics[width=\fWidD]{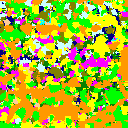}&
\includegraphics[width=\fWidD]{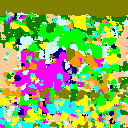}
\end{tabular}
}

\begin{tabular}{c|cccc}
    \hline
   &SAN-img & SAN-flow & +Cond. & DivA \\
    \hline
bIoU & 9.24 & 22.05 & 59.87 & \textbf{64.16} \\
    \hline
\end{tabular}
\caption{\sl\small {\bf Results on Movi-Solid.} DivA outputs compelling segmentation, while SAN struggles with object boundaries. Adding a conditional decoder to SAN improves boundary alignment but is still vulnerable to over-segmentation. Naively applying SAN for image reconstruction (SAN-img) yields non-informative results, emphasizing the role of motion cues in object discovery. }
\label{fig:movi}
\vspace*{-0.1in}
\end{figure}

\textbf{Multi-object segmentation on FBMS-59. } FBMS-59 dataset contains 59 videos ranging from 19 to 800 frames. In the test set, 69 moving objects are labeled in 30 videos. We train with $n=4$ to extract \emph{multiple} independently moving objects. At test time, we vary $n$ from 3 to 6 without re-training, and report bootstrapping IoU. We test the model on optical flow computed with $\delta t = 1$ and $\delta t = 2$, and keep the spatial resolution to $128\times128$ in both training and testing. 

\begin{figure}[t]
\renewcommand{\arraystretch}{0.2} 
\def\fWidD{0.086\textwidth}
\footnotesize
\hspace*{-1mm}
\begin{tabular}[0mm]{c@{\hskip 0.01in}c@{\hskip 0.01in}c@{\hskip 0.01in}c@{\hskip 0.01in}c}
Image&Flow&2 slots&3 slots&4 slots\\
\includegraphics[height=\fWidD]{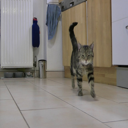}&
\includegraphics[height=\fWidD]{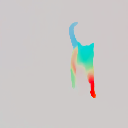}&
\includegraphics[height=\fWidD]{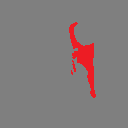} &
\includegraphics[height=\fWidD]{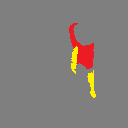} &
\includegraphics[height=\fWidD]{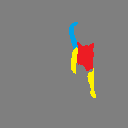}\\
\includegraphics[height=\fWidD]{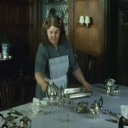}&
\includegraphics[height=\fWidD]{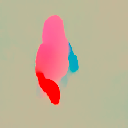}&
\includegraphics[height=\fWidD]{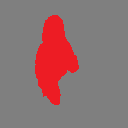} &
\includegraphics[height=\fWidD]{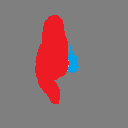} &
\includegraphics[height=\fWidD]{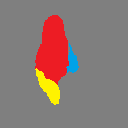}
\end{tabular}

\caption{\sl\small {\bf Varying the number of slots changes the granularity of segmentation.} All the above results are obtained from the same model by only varying the number of slots at inference time, without re-training. This gives users additional control over the granularity of segmentation.}
\label{fig:granularity}
\end{figure}

Tab.\ref{tab:fbms} summarizes the results. The conditional decoder improves segmentation accuracy substantially, and the adversarial training further improves accuracy by 3.9\% in the best-performing case. We notice that $n=4$ is a sweet spot for the dataset, due to the dataset's characteristic of having less than 3 moving objects in most sequences. Fig.~\ref{fig:fbms} shows qualitative examples.
For many objects segmented, we observe a change in the level of granularity of the segmentation mask when varying $n$. Two examples are given in Fig.~\ref{fig:granularity}. Although depending on the annotation, both cases may be considered as over-segmenting the object in quanlitative evaluations and degrade IoU, we argue that such a feature is desirable. As DivA can conveniently vary $n$ without re-training, it reveals DivA's potential in developing adaptive, interactive, or hierarchical segmentation in future work. 

\begin{table}[t]
\footnotesize
  \centering
  \hspace*{-0.1in}
  \begin{tabular}{l|c|c|c|c|c}
    Method&Resolution&Multi&DAVIS&ST&FPS\\
      \hline\multicolumn{6}{c}{Unsupervised Single-frame Method}\\
      \hline
    {CIS \cite{yang2019unsupervised} (4)}&$128\times224$&N&59.2&45.6&10\\
    {CIS (4)+CRF}&$128\times224$&N&71.5&62.0&0.09 \\
    {MoSeg \cite{yang2021self}}&$128\times224$&N&65.6&-&78 \\
    {MoSeg (4)}&$128\times224$&N&68.3&58.6&20 \\
    {EM \cite{meunier2022driven}}&$128\times224$&Y*&69.3&60.4&21 \\
    \hline
    {DivA-ZeroShot}&$128\times224$&Y&66.0&57.5&64\\
    {DivA}&$128\times128$&Y&68.6&60.3&\textbf{104}\\
    {DivA}&$128\times224$&Y&70.8&60.1&64 \\
    {DivA-R}&$128\times224$&Y&71.0&60.9&66 \\
    {DivA(4)}&$128\times224$&Y&\textbf{72.4}&\textbf{64.6}&16 \\  
        \hline\multicolumn{6}{c}{With Additional Supervision}\\
      \hline
          {FSEG \cite{fusionseg}}&Full&N&70.7&61.4&5 \\
       {OCLR \cite{xie2022segmenting}}&$128\times224$&Y&72.1&67.6&- \\
        {ARP \cite{koh2017primary}}&Full&N&76.2&57.2&0.015 \\
    {DyStab \cite{yang2021dystab}}&Full&N&80.0&73.2&0.09 \\
          \hline\multicolumn{6}{c}{Video Batch Method}\\
      \hline
      {ELM \cite{lao2018extending}}&Full&Y&61.8&-&-\\
        {IKE \cite{haller2021iterative}}&$224\times416$&N&70.8&66.4&- \\
          {USTN \cite{Meunier_2023_CVPR}}&$128\times224$&Y*&73.2&55.0&- \\
      {DS \cite{ye2022sprites}}&Full&Y&79.1&72.1&-

    \end{tabular}\vspace*{-0.1in}
  \caption{\sl\small{\bf Binary segmentation results.} Although not exploiting the prior of a dominant ``foreground object'', DivA still achieves compelling accuracy on binary segmentation, with better inference speed. DivA closes the gap between single-frame unsupervised methods to methods using additional supervision, pre-trained features, or video batch processing exploiting temporal consistency (not applicable to embodied systems). Multi: method generalizes to multi-object segmentation; DivA-R: recursive implementation.}
    \label{tab:results}
\vspace*{-0.1in}
\end{table}

\textbf{Multi-object segmentation on Movi-Solid.} 
Movi-Solid\footnote{The ground-truth for Movi-Solid is not publicly available. Quantitative results are reported on a subset of the dataset that we manually labeled.}  \cite{singh2022simple} is a realistic synthetic dataset containing 9000 sequences featuring various, unidentified numbers of objects. Most objects exhibit motion relative to the dynamic background. We train DivA on $128 \times 128$ resolution with 4 slots. At test time, we increase the number of slots to $n=10$. As in Fig.~\ref{fig:movi}, DivA yields compelling segmentation results. The SAN model, although trained with the same reconstruction loss, fails to capture object boundaries. When equipped with a conditional decoder, SAN yields better mask alignment with object boundaries but tends to over-segment due to the absence of an information separation criterion. For completeness, we also train SAN on image reconstruction denoted as ``SAN-img''. Given the complexity of nuisance in the dataset, it fails to yield meaningful segmentation. This underscores the significance of using motion as a robust cue for unsupervised object discovery.

\textbf{Binary segmentation. }
We also present results on binary segmentation datasets, a special case where $n=2$, due to the scarcity of benchmarks and baselines for multi-object motion segmentation: \textbf{DAVIS2016} \cite{perazzi2016benchmark} consists of 50 videos, each ranging from 25 to 100 frames containing one primary moving object that is annotated. \textbf{SegtrackV2} \cite{tsai2012motion} has 14 video sequences with a total of 947 frames. Objects having apparent motion relative to the background are annotated.

We compare with:
\textbf{CIS} \cite{yang2019unsupervised} uses a conventional binary segmentation architecture. It enforces information separation by minimizing mutual inpainting accuracy on the regions.
\textbf{Moseg} \cite{yang2021self} adopts SAN for iterative binding to the flow field, fixes the number of slots to 2, and learns slot initialization instead of using random Gaussian.
\textbf{EM} \cite{meunier2022driven} pre-defines a class of target motion patterns ({\em e.g.,} affine), and alternately updates segmentation and corresponding motion patterns during training. We select them as baselines since they 1) use single-frame without exploiting cross-frame data association; 2) do not rely on the presence of a dominant ``foreground object''. 
DivA is chosen to be efficient and flexible, leading to 1), and not biased towards purposefully framed videos, leading to 2). For completeness, we also include paragon methods:  \textbf{IKE}~\cite{haller2021iterative}, \textbf{USTN}~\cite{Meunier_2023_CVPR}, \textbf{OCLR}~\cite{xie2022segmenting} and \textbf{DS}~\cite{ye2022sprites} that operate on image batches, and \textbf{FSEG}~\cite{fusionseg}, \textbf{ARP}~\cite{koh2017primary}, and \textbf{DyStab}~\cite{yang2021dystab} that require additional supervision.

\begin{figure}[t]
\renewcommand{\arraystretch}{0.2} 
\def\figd{figures/flipped}
\def\fWidD{0.085\textwidth}
\footnotesize
\hspace*{-5mm}
\begin{tabular}[0mm]{c@{\hskip 0.01in}c@{\hskip 0.01in}c@{\hskip 0.01in}c@{\hskip 0.01in}c}
&Input Flow&MoSeg&DivA
\\
\smash{\rotatebox{90}{\quad Original}}&\includegraphics[height=\fWidD]{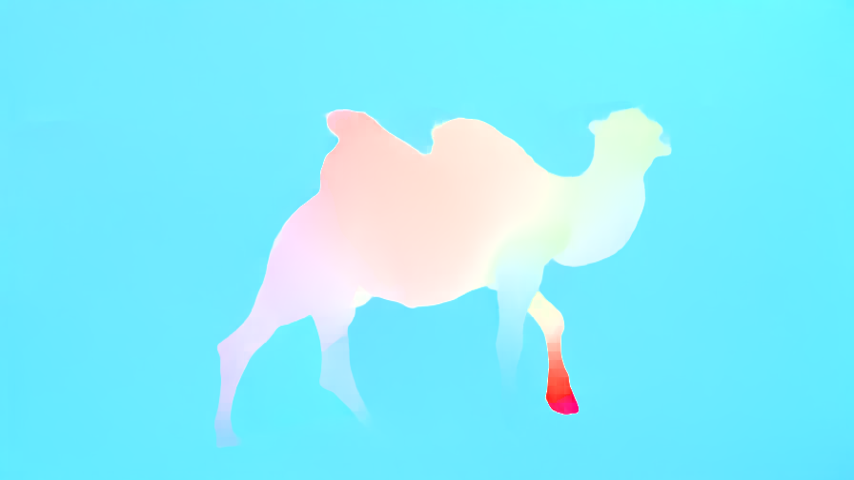} & \includegraphics[height=\fWidD]{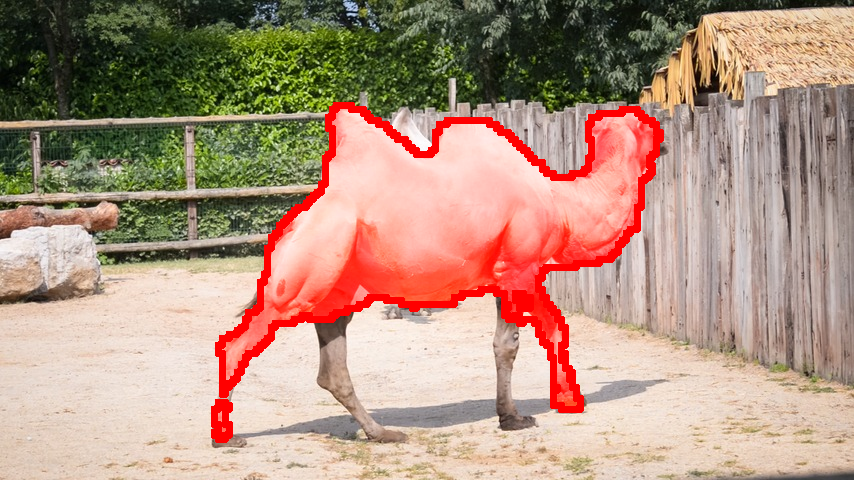} &
\includegraphics[height=\fWidD]{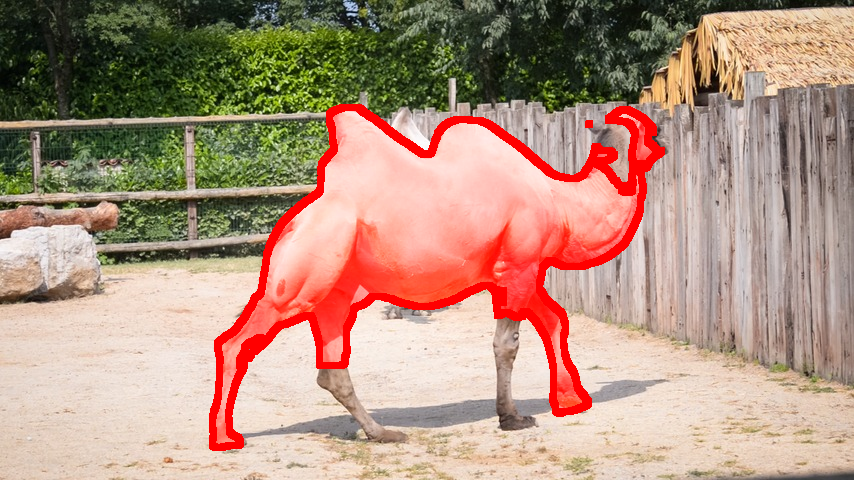} 
\\
\smash{\rotatebox{90}{\quad Flipped}}&\includegraphics[height=\fWidD]{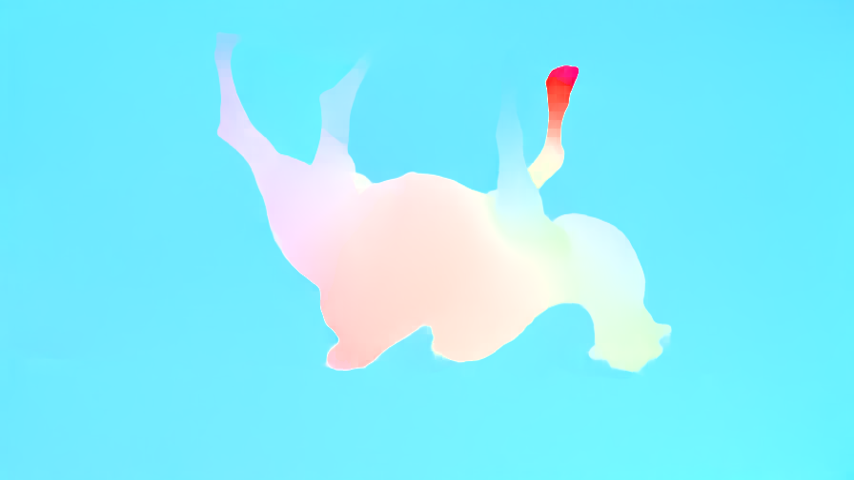} & \includegraphics[height=\fWidD]{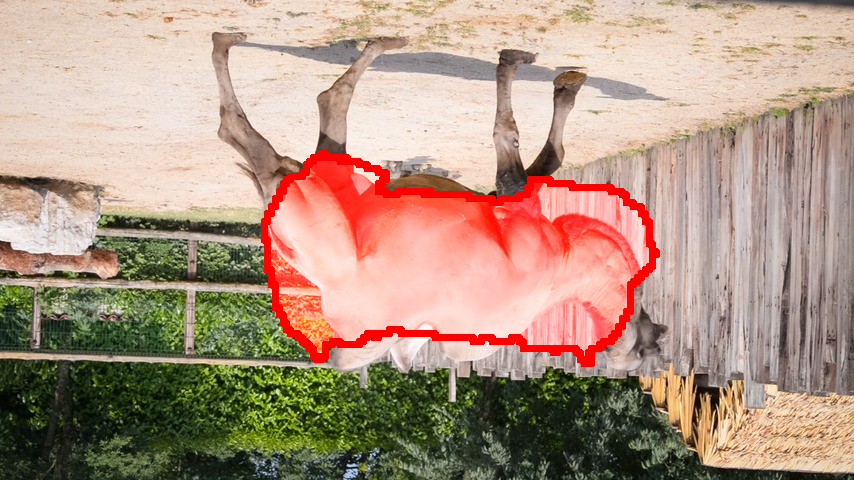} &
\includegraphics[height=\fWidD]{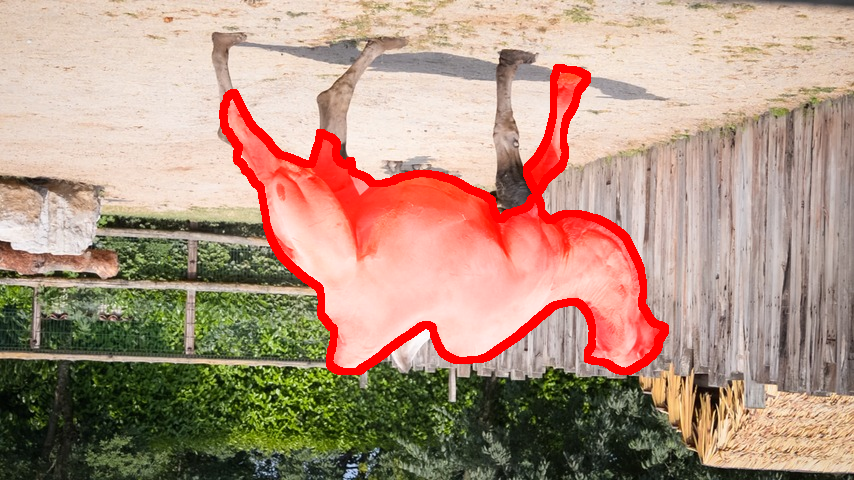}
\end{tabular}
\caption{\sl\small {\bf Conditional decoder is less vulnerable to overfitting.} Reconstructing complex input from slots forces the decoder to overfit to seen data. Simply flipping the input image and flow drastically decreases segmentation accuracy. By introducing the conditional decoder, DivA is less vulnerable to overfitting.}
\label{fig:inverted}\vspace*{-0.1in}
\end{figure}

Results are summarized in
Tab.\ref{tab:results}. Our best result outperforms the best-performing baseline EM by 4.5\% and 6.9\%, respectively. We measure the run time of one single forward pass of the model. Our fastest setting using $128\times128$ resolution reaches 104 frames per second. Note that the speed is reported excluding optical flow estimation cost following the baseline methods. When combined with RAFT optical flow (40 fps on $128\times128$ resolution), the fastest version of DivA runs at 28 fps including optical, making it ideal to be deployed to embodied systems. In addition to running on consecutive frames, CIS and MoSeg merge segmentation results on optical mapping from the reference frame to 4 adjacent temporal frames, marked by ``(4)'' in the table. The same protocol improves DivA on both datasets. Compared to models using additional supervision, our best performance exceeds FSEG (supervised) and OCLR (supervised, trained on synthetic data), and closes the gap with the best-performing method DyStab to 12\%. Inspired by \cite{elsayedsavi++}, we test DivA by recursively inheriting slots instead of randomly initializing them, and reducing the number of iterations from 3 to 1. Both segmentation accuracy and runtime speed improve marginally, with reduced label-flipping. Unlike MoSeg and DyStab which employ additional regularizations, our recursive model exhibits temporal consistency without modification to the training pipeline.

\textbf{Zero-shot generalization.} Note that Moseg and CIS require training on target sequences. To demonstrate DivA's zero-shot transfer capabilities, we train only on YouTube-VOS \cite{vos2018}, and test on DAVIS without any fine-tuning (including training set). The results are labeled as 'DivA-ZeroShot'. Despite this, the performance of DivA still outperforms the baselines, showcasing the robustness of the method. Furthermore, we compare with MoSeg which also uses slot attention. Without the conditional decoder, due to the limited capacity of the slots, MoSeg relies on the decoder to reconstruct fine structures of the input flow, forcing the decoder to memorize shape information. Together with learned slot initializations, the model is subject to overfitting. Fig.~\ref{fig:inverted} shows an example, where simply flipping the input leads to performance degradation. Conditioning the decoder on the image frees it from memorizing the flow to be reconstructed, thus making the model less vulnerable to overfitting.

\textbf{Different optical flow.}
It is well-established \cite{julesz1971foundations} that motion estimation, such as optical flow, can be inferred through optimization without the need for induction or human supervision \cite{horn1981determining}. However, to follow the baselines \cite{yang2019unsupervised, yang2021self, bao2022discovering, bao2023object, meunier2022driven}, we use RAFT optical flow, which is trained on synthetic datasets with supervision, as an off-the-shelf component. As a precautionary measure, we also test SMURF \cite{stone2021smurf}, an optical flow method that does not require supervision, to ensure that our approach does not inadvertently inherit bias from the supervised training of optical flow. The results, presented in Tab.~\ref{tab:smurf}, show that performance on the DAVIS dataset remains consistent with the baseline methods, demonstrating that our model is robust to different optical flow techniques and training sources.

\def\figd{figures/n_slots}

\begin{table}[t]
\footnotesize
\setlength\tabcolsep{5pt}
  \centering
\begin{center}
\begin{tabular}{c|ccc}
    \hline
SMURF Source Dataset & Chairs \cite{dosovitskiy2015flownet} & KITTI \cite{Geiger2012CVPR} & Sintel \cite{Butler:ECCV:2012} \\
    \hline
J-score on DAVIS & 66.4 & 65.4 & 67.5 \\
    \hline
\end{tabular}
\vspace{-1mm}
\end{center}
\vspace*{-0.1in}
  \caption{\sl\small{\bf Segmentation results using self-supervised optical flow trained on different sources.}  As a sanity check, we evaluate DivA using the self-supervised optical flow method SMURF \cite{stone2021smurf}, which is trained on different source datasets. The segmentation results remain competitive and comparable to the baseline methods.}
    \label{tab:smurf}
    \vspace{-2mm}
\end{table}

\textbf{Contrastive learning on bootstrapped objects. }
While DivA operates on videos by utilizing motion, we want to showcase that it can facilitate the training of a versatile vision encoder for 2D static images. State-of-the-art self-supervised, contrastive learning methods \cite{caron2021emerging, chen2020simple, he2020momentum}  often rely on training data containing "dominant objects," making it challenging to apply them directly to generic video data. Meanwhile, DivA autonomously discovers objects, making it a valuable complement to contrastive learning. To demonstrate this, we directly apply pre-trained DivA to YouTube-VOS \cite{vos2018} and MOSE \cite{MOSE} datasets, the combination of which comprises 4939 video clips containing various numbers of objects with unknown semantic categories. Although neither dataset is curated specifically for motion segmentation, DivA successfully bootstraps a large number of (in total $\sim$200k) object candidates (Fig.~\ref{fig:dino}) based solely on motion cues. We then train a DINO model on these bootstrapped candidates, following the original hyperparameters except for a batch size of 160, constrained by GPU memory. 

We evaluate the trained models by linear probing on ImageNet \cite{deng2009imagenet} and dense object tracking on DAVIS2017 \cite{Pont-Tuset_arXiv_2017}, adhering to the default protocols following DINO without fine-tuning the backbone (zero-shot). Compared to models trained on original video frames, DivA narrows the performance gap to training on ImageNet by as much as 30.2\%, all with only a mere 1.5\% increase in total training time and no human intervention. Notably, on less than 5000 videos, where the object classes are neither balanced nor aligned with ImageNet categories, our approach yields an ImageNet top-1 accuracy of 43.6\%. This demonstrates the effectiveness of bootstrapping by DivA, and opens up the potential of automating training visual foundation models on large-scale, generic videos, without relying on purposefully curated data (e.g. \cite{oquab2023dinov2}). We expect that scaling up the dataset will enhance performance and eventually surpass ImageNet-based training. Due to computational constraints, however, this broader investigation is deferred to future work.

\section{Discussion}
\label{sec:discussion}

When discussing the scale of supervision, it is crucial to recognize that human supervision extends beyond explicit labels. For instance, an implicit image selection bias is often unintentionally embedded in Internet-scraped images due to a strong tendency for people to upload photos framed into ``shots''. Consequently, ``unsupervised" or ``self-supervised" approaches that exclude labels may still inherit substantial human cognitive priors.
While we acknowledge that creating a model completely free of human bias may be impractical given the data we train on is ultimately sourced by humans, this study, though idealistic, provides a closer approximation to the {\em ab-ovo} development of visual perception. By bootstrapping ``detached objects'' solely by the natural cue of motion, we investigate how object perception can emerge without symbolic annotations or viewpoint selection bias.

The resulting approach, DivA, is data-driven, trains on unrestricted videos without labels or pre-trained features, and is the first real-time multi-region motion segmentation model that works on instantaneous motion handling multiple and unknown numbers of objects. These properties are shared with biological systems that have similarly evolved without explicit supervision, and similarly DivA is fast enough to be employed in a closed loop in embodied settings. 


DivA enjoys certain properties not observed in existing methods: {\em First}, it does not require specifying the number of objects, and allows for their upper bound to change at test time.  
{\em Second}, and DivA can be used recursively to reduce label flipping by simply using slots from previous frames as initialization for incoming frames. Other methods for doing so  \cite{elsayedsavi++} require manual input and modification of the training pipeline.
{\em Third}, DivA can be trained on images of a certain resolution and then used on different resolutions at inference time. We have trained on $128\times 128$ to spare GPU memory and tested on $128\times 224$ for accuracy, and vice-versa one can train on larger images and use at reduced resolution for speed. 
{\em Fourth}, DivA is fast.
{\em Finally}, DivA can upsample segmentation masks to high resolution, without post-processing ({\em e.g.} CRF) to refine the results.

\begin{table}[t]
\setlength\tabcolsep{5pt}
\footnotesize
  \centering
  \begin{tabular}{l|c|c|c|c|c}
    \multirow{2}{*}{\diagbox[height=2\line,outerleftsep=25pt, width=17pt]{\hspace*{-1.2cm}Pre-train\quad Downstream\\\hspace*{-1.2cm}data source\quad accuracy}{}}&\multicolumn{2}{c|}{Classification}&\multicolumn{3}{c}{Video object segmentation}\\
\cline{2-6}
&Top-1&Top-5&F\&J&J-mean&F-mean\\
      \hline
Videos&37.9&60.4&54.1&52.3&55.8\\
Videos + DivA&\textbf{43.6}&\textbf{66.3}&\textbf{55.7}&\textbf{54.2}&\textbf{57.2}\\        \hline\multicolumn{6}{c}{Paragon}\\
      \hline
ImageNet&73.0&90.8&60.7&58.6&63.8\\
\hline
\hline
Relative gap reduced&16.1\%&19.3\%&24.2\%&30.2\%&17.5\%\\
\end{tabular}\vspace*{-0.1in}
  \caption{\sl\small{\bf Objects discovered by DivA facilitates contrastive learning.}  Contrastive learning methods often rely on object-centric training data, making it challenging to apply them directly to generic video data. Meanwhile, DivA autonomously discovers objects, making it a valuable complement to contrastive learning. By using DivA to bootstrap objects from YouTube-VOS and MOSE, we narrow the performance gap to ImageNet training by up to 30.2\%.}
    \label{tab:dino}
\vspace*{-0.1in}
\end{table}

DivA has several limitations. While we have tested DivA on a handful of objects, scaling to hundreds or more slots is yet untested. Common failure modes of DivA are analyzed in the Supp. Mat. With more powerful foundation models, the role of motion may become diminished, but the question of how foundation models are trained, remains. Since our goal is to understand the emergence of semantics, we keep DivA free of semantic annotation or pre-trained features. 

There are many variants possible for the architecture in Fig.~\ref{fig:schematic}, including using Transformers (e.g. \cite{singh2021illiterate, singh2022simple}) in lieu of the conditional prior network. Even though the amphibian visual system requires objects to move to spot them, primate vision can easily detect and describe objects in static images, thus, a bootstrapping mechanism can be built from DivA. Moreover, varying the number of slots changes the level of granularity of the slots (Fig.~\ref{fig:granularity}), which leads to the natural question of how to extend DivA to hierarchical partitions. 

{
    \small
    \bibliographystyle{ieeenat_fullname}
\bibliography{references}
}

\newpage
\appendix
\onecolumn

\section{Implementation Details}\label{sec:implementation}
\begin{figure}[t]
\centering
\includegraphics[width=0.4\textwidth]{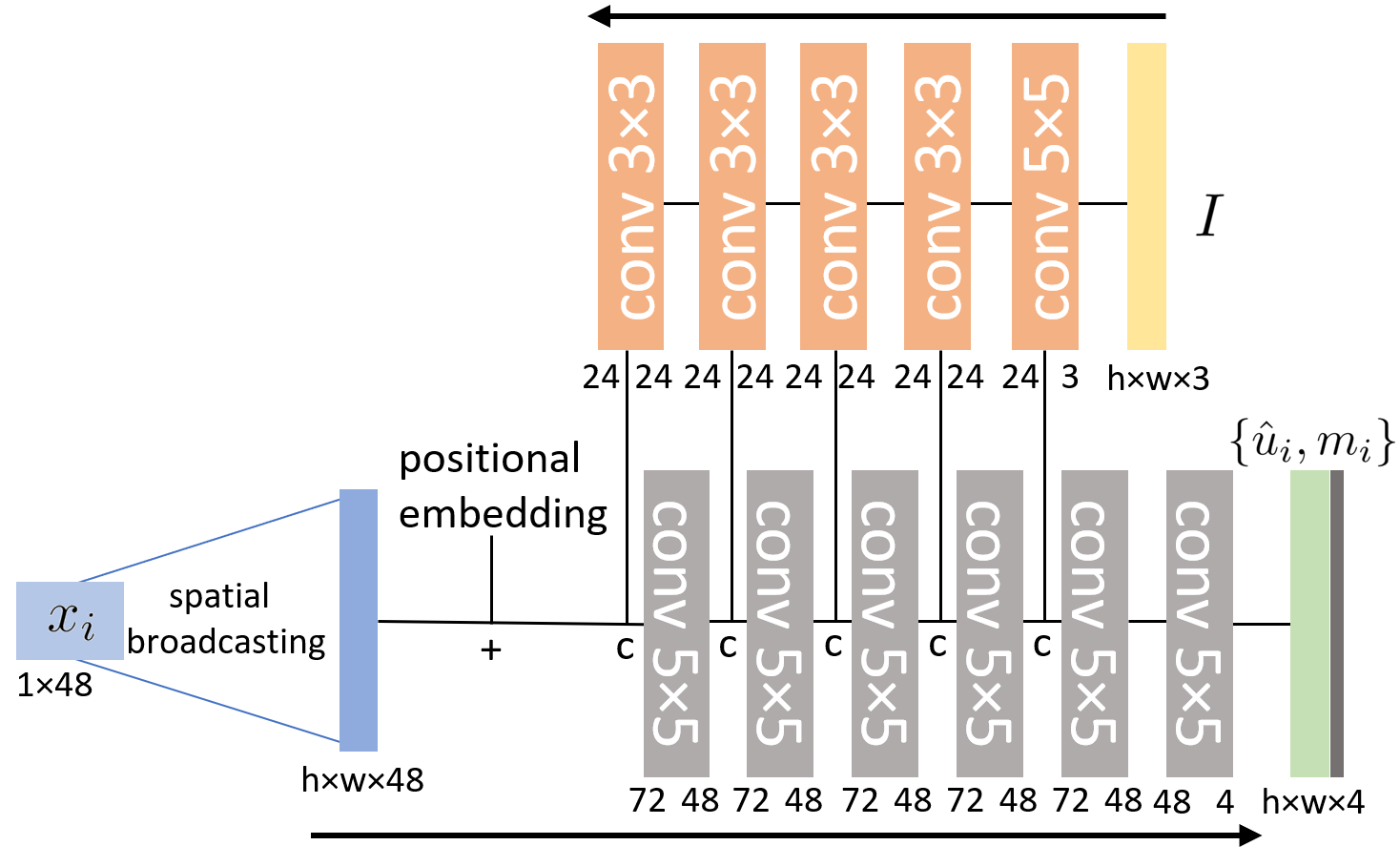}
    \caption{\sl\small {\bf Architecture of the conditional decoder.}}
    \label{fig:decoder}
    \vspace{-0.2cm}
\end{figure}

\textbf{Training and testing. } We apply RAFT \cite{teed2020raft} for optical flow
following \cite{yang2021self,meunier2022driven}. For image sequence $\{I_t\}$, RAFT computes optical flow from $I_t$ to $I_{t+\delta t}$, where $\delta t$ is sampled from -2 to 2. Optical flow then downsampled together with the reference image before feeding to the network. During training, we warm up our model on DAVIS2016 dataset, setting $n=4$. During this warm-up stage, we set $\lambda=0$ so that the model can learn reconstruction in the initial stage without the interference of an adversarial decoder. After the warm-up, we train the model on each particular dataset, with $n=2$ on DAVIS2016 and SegtrackV2, and $n=4$ on FBMS-59. Referring to the empirical evidence on the diagnostic dataset, We set $\lambda=0.03$, and decrease it to $\lambda-0.01$ towards the end of the training. We use a batch size of 32 and set the spatial resolution of model input to be $128\times128$ when $n=4$ and $128\times224$ when $n=2$, tailored by the GPU's memory constraint.

At test time, we keep $n=2$ on DAVIS2016 and SegtrackV2, and vary $n$ on FBMS-59. On binary segmentation, since the model is trained without the notion of foreground and background, we follow the baselines to match the ground truth with the most likely segmentation mask and compute intersection-over-union (IoU) to measure the segmentation accuracy. We upsample the segmentation output to the dataset's original resolution for evaluation. Unlike baseline methods that upsample low-resolution segmentation masks by interpolation, our generative model reconstructs flow from each slot, so we can refine the segmentation outputs by upsampling $\{\hat{u}_i\}$'s to the full resolution, then refine the segmentation boundaries by $\argmin_i|u-\hat{u}_i|_2$. This practice creates negligible computational overhead (0.00015s) and empirically improves segmentation accuracy. On multi-object bootstrapping, we follow the bootstrapping IoU defined in Sect.~\ref{sec:results} and evaluate the accuracy per instance. Note that this is different from the baseline methods that merge all instances into a single foreground for evaluation.

\textbf{Data Normalization. }
DivA takes optical flow and image as input. Since there is no restrictions on the magnitude of the flow, we normalize the input RGB flow to zero-mean and range in [-1,1].\footnote{One may convert the 2-channel flow field to an RGB image by a standard color-wheel, following CIS \cite{yang2019unsupervised} and MoSeg \cite{yang2021self}. Both normalizations yield similar performance in our experiments.} We notice that normalizing the flow to zero-mean, which eliminates a global translation in the motion field, leads to more stable training. The input image is also normalized to within [-1,1], similar to SAN \cite{locatello2020object}. 

\textbf{Training Parameters. } For $w$ and $\theta$ we train with two separate ADAM optimizers to perform alternating optimization. In the warm-up stage, $w$ is trained with $\lambda=0$ (no adversarial loss). Note that even if the adversarial loss is not applied, we still update $\theta$ at this stage, so the adversarial decoder still learns to reconstruct the flow on the whole image domain from each slot. Both optimizers use an initial learning rate of $8e^{-4}$ with linear learning rate decay. On diagnostic data, we train for 1000 epochs. On real-world data, the network first warm-ups for 200 epochs on DAVIS with $n=4$ and spatial resolution $128\times 128$, then train on each dataset under settings specified in Sect. 3.2 in the main paper for different experiments. This warm-up step provides a stable initialization to the network, especially when the target dataset is small (e.g. SegTrackV2). On DAVIS and SegTrackV2 the network is trained for 500 epochs, and on FBMS-59 the network is trained for 200 epochs since the dataset is larger.

\section{Additional Qualitative Tests}

\def\figd{figures/failure_cases}
\def\fWidD{0.12\textwidth}
\begin{figure}[t]
\renewcommand{\arraystretch}{0.2} 
\centering
{\footnotesize
\hspace*{-0.1in}
\begin{tabular}[0mm]{c@{\hskip 0.01in}c@{\hskip 0.01in}c@{\hskip 0.01in}c}
Image&Flow&Reconstruction&Segmentation \\
\includegraphics[width=\fWidD]{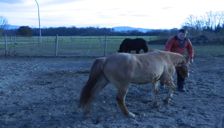}&
\includegraphics[width=\fWidD]{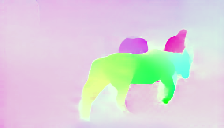}&
\includegraphics[width=\fWidD]{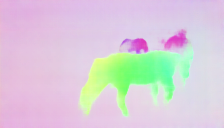} &
\includegraphics[width=\fWidD]{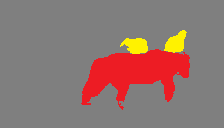}\\
\includegraphics[width=\fWidD]{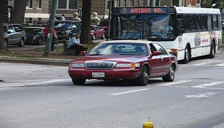}&
\includegraphics[width=\fWidD]{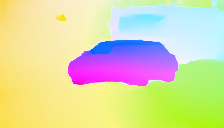}&
\includegraphics[width=\fWidD]{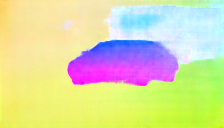} &
\includegraphics[width=\fWidD]{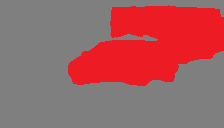}
\end{tabular}

\caption{\sl\small {\bf Failure modes.} DivA is trained using the CIS principle, which assumes objects move independently. Consequently, two objects moving in the same way may share high mutual information, e.g., even though the car and the bus are different detached objects, if their motions are highly correlated in the current video, they are seen as one by DivA (of course, as time goes by, their motions will diverge, allowing DivA to correctly separate them).}
\label{fig:failurecases}
}
\end{figure}

\def\figd{figures/suppmat/upsample}
\def\fWidD{0.16\textwidth}
\begin{figure}[t]
\centering
\hspace*{-3mm}
{\footnotesize
\begin{tabular}[0mm]{c@{\hskip 0.01in}c@{\hskip 0.01in}c}
Optical Flow&Output Mask&Ours\\
\includegraphics[width=\fWidD]{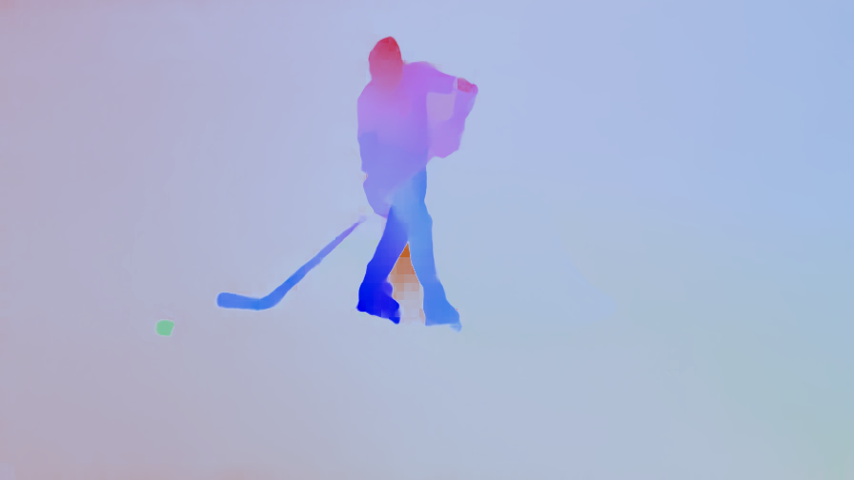}&
\includegraphics[width=\fWidD]{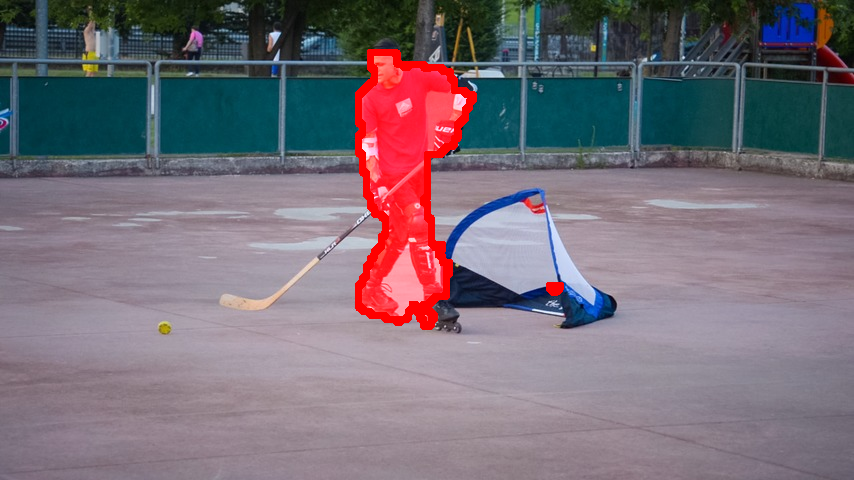}&
\includegraphics[width=\fWidD]{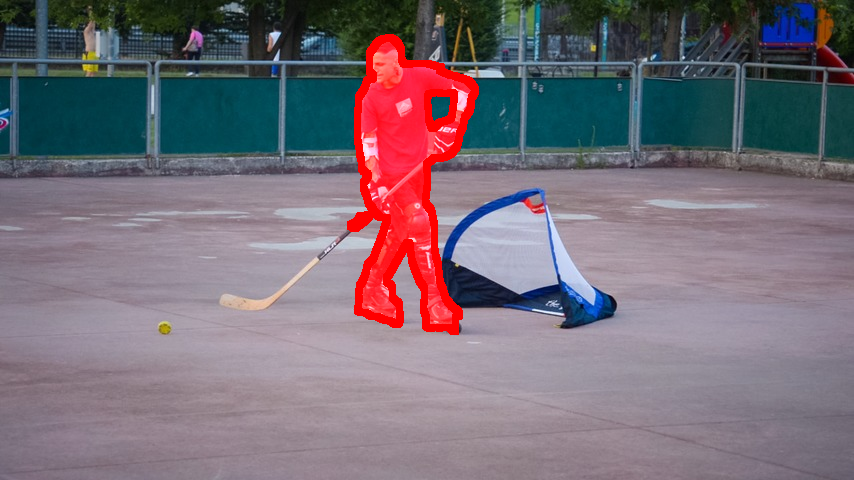} \\
\includegraphics[width=\fWidD]{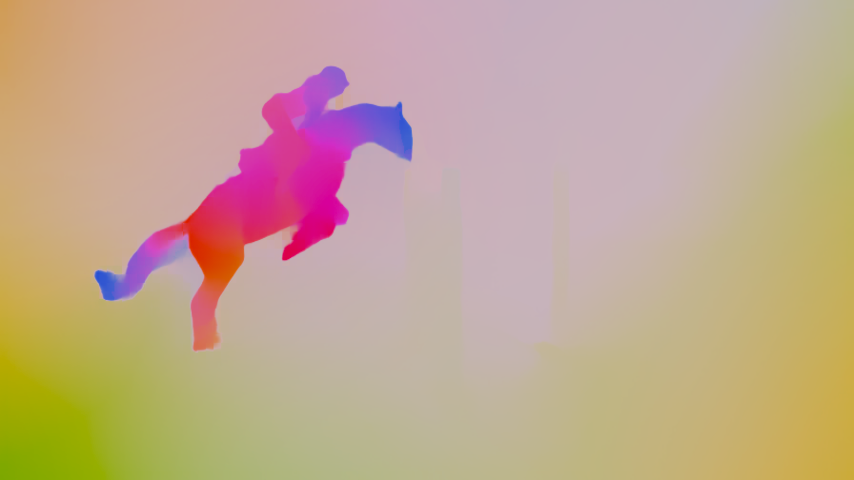}&
\includegraphics[width=\fWidD]{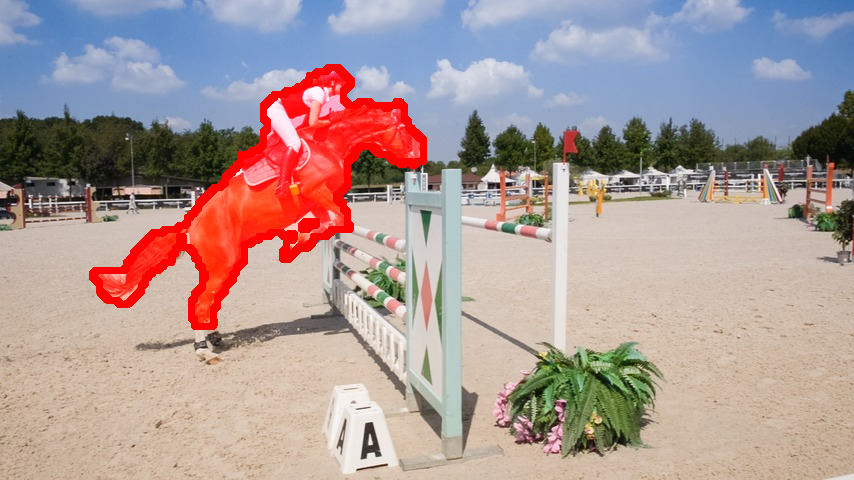}&
\includegraphics[width=\fWidD]{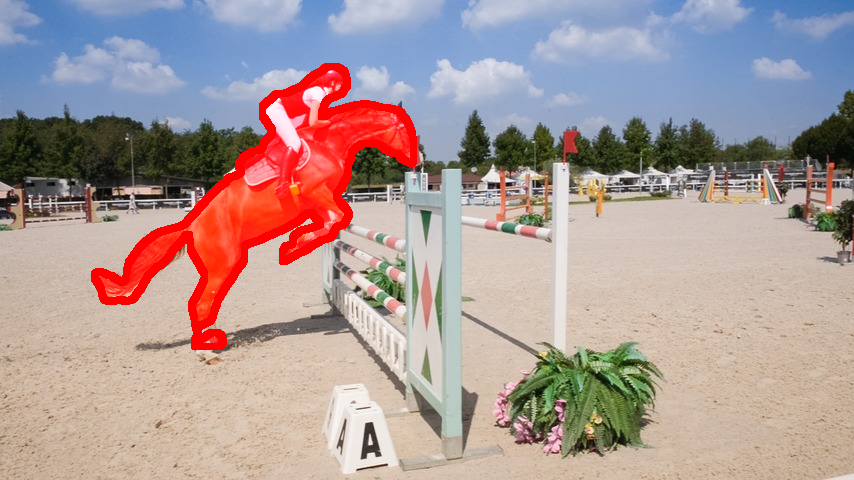} 
\end{tabular}\\
\includegraphics[width=.24\textwidth]{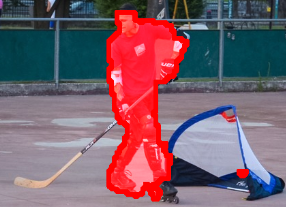}\includegraphics[width=.233\textwidth]{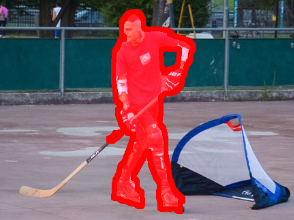}   
}
\caption{\sl\small {\bf Upsampling segmentation masks to full resolution.} Better viewed zoomed-in. Our generative model allows upsampling segmentation masks to full resolution without losing details of object boundaries. Details for the upper panel are shown side-by-side in the last row.}
\label{fig:resolution}
\end{figure}

\textbf{Upsampling segmentation masks to full resolution. }
Many motion segmentation techniques rely on low-resolution input data \cite{yang2019unsupervised, yang2021self, meunier2022driven}, and output segmentation masks are then upsampled through interpolation to achieve full resolution. However, this interpolation process leads to a loss of detail, especially along object boundaries. In contrast, the DivA generative model offers an alternative approach to upsampling that involves comparing upsampled flow reconstruction with the full-resolution optical flow by $\argmin_i|u-\hat{u}_i|_2$, thereby refining the segmentation boundaries without significantly increasing computational overhead. This allows for the segmentation masks that match the optical flow at full resolution, producing smoother segmentation boundaries that align well with the moving objects in the image. Fig.~\ref{fig:resolution} provides representative samples of the outcome of this upsampling procedure  (better visible after zooming-in). Compared with directly upsampling the output mask (second column), our approach (third column) captures fine details of object boundaries without the artifacts that are typically associated with upsampling.

\textbf{Conditional Decoder. }
The DivA model employs a conditional decoder that utilizes an image as a prior to reconstructing optical flow from slots. Two examples of the reconstructed optical flow are presented in Fig.~\ref{fig:cond}. In these examples, the input flow misses the fine structure of the object due to the assumptions underlying flow estimation, which causes artifact in the resulting flow estimate. Despite the flaws of the input flow, the reconstruction through the conditional decoder correctly captures the object boundaries. This illustrates the fact that the decoder learns to reconstruct the flow by leveraging photometric information such as high-contrast edges, their shape, and more general photometric structures in the image. In the example in the figure, the conditional prior helps the model troubleshoot the ``mono-leg'' in the middle column. This opens up the potential of bootstrapping objects in static images by motion directly from the DivA model.

\textbf{Different values of $k$. }
We provide an ablation study (Fig.~\ref{fig:k})on the size of each slot below. While $k=48$ and $k=64$ achieve similar reconstruction loss, $k=48$ exhibits a higher adversarial loss, indicating that while both encode motion within the mask effectively, $k=64$ encodes external motion better (overfitting). $k=48$ yields satisfactory results empirically.

\def\figd{figures/rebuttal}
\def\fWidD{0.6\textwidth}

\begin{figure}[t]
\centering
\includegraphics[width=\fWidD]{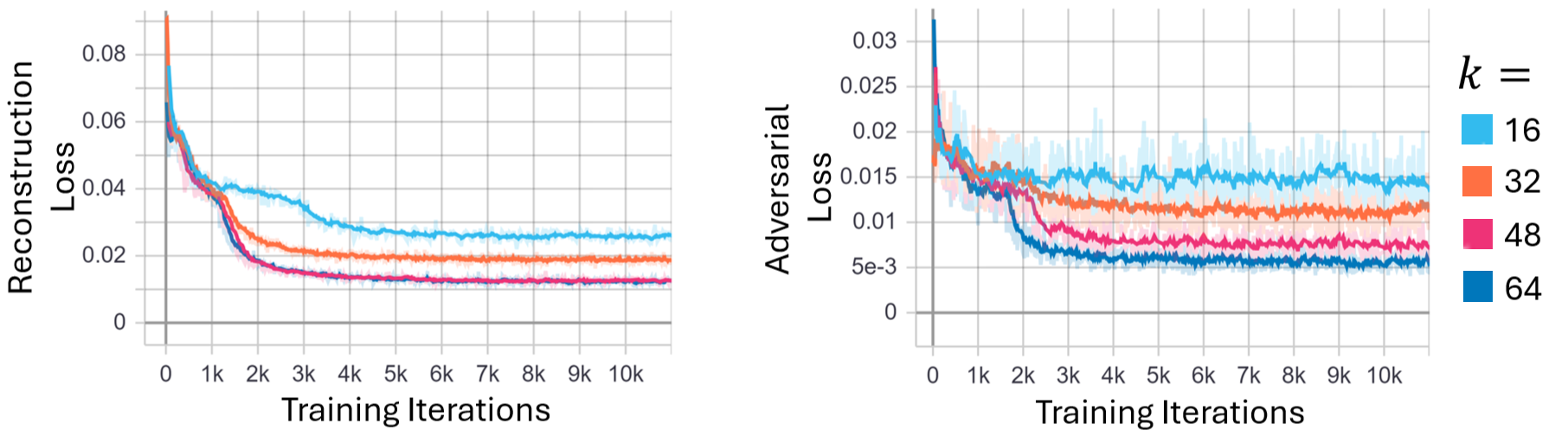}
\caption{\sl\small {\bf Different slot sizes.} While both \( k = 48 \) and \( k = 64 \) achieve similar reconstruction losses, \( k = 48 \) results in a higher adversarial loss. This suggests that while both values effectively encode motion within the mask, \( k = 64 \) better captures external motion, which may lead to overfitting. Based on these observations, \( k = 48 \) provides satisfactory results empirically.
}
\label{fig:k}
\end{figure}

\def\figd{figures/suppmat/decoder}
\def\fWidD{0.11\textwidth}
\begin{figure}[t]
\centering
\hspace*{-0mm}
{\footnotesize
\begin{tabular}[0mm]{c@{\hskip 0.1in}c@{\hskip 0.1in}c}
Input Image&Input Flow&Reconstructed Flow\\
\includegraphics[width=\fWidD]{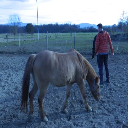}&
\includegraphics[width=\fWidD]{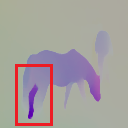}&
\includegraphics[width=\fWidD]{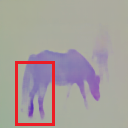} \\
\includegraphics[width=\fWidD]{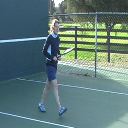}&
\includegraphics[width=\fWidD]{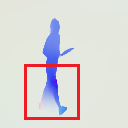}&
\includegraphics[width=\fWidD]{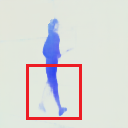} \\
\end{tabular}
}
\caption{\sl\small {\bf Conditional decoder learns image priors.} Despite imperfections in the input optical flow within the highlighted region, our reconstructed flow exhibits a satisfactory alignment with object boundaries. This finding indicates that the conditional decoder effectively employs image priors for flow reconstruction.}
\label{fig:cond}
\end{figure}

\textbf{Examples for Diagnostic Data. }
Fig.~\ref{fig:diagnostic} shows representative examples in the diagnostic data. We utilize objects and corresponding segmentation masks extracted from the DAVIS dataset and merged them with arbitrary background images. Objects (including the background) are assigned motion patterns that are statistically independent. An image may contain $n = 2,3,4$ independent motion patterns and the network is unaware of the exact number. We use complex natural images so that the network cannot overfit to simply segmenting RGB images. 

On our diagnostic dataset, the vanilla slot attention model fails to output accurate segmentation. With the introduction of a conditional decoder, masks align better to object boundaries, but multiple slots may map to the same motion. The full DivA model further promotes information separation between slots, especially when the number of slots exceeded the number of independent motions, resulting in more accurate segmentation outputs.

\section{Additional Discussion}
\def\figd{figures/adv}
\def\fWidD{0.08\textwidth}
\begin{figure}[t]
\renewcommand{\arraystretch}{0.2} 
\centering
\hspace*{-5mm}
{\footnotesize
\begin{tabular}[0mm]{c@{\hskip 0.01in}c@{\hskip 0.01in}c@{\hskip 0.01in}c@{\hskip 0.01in}c@{\hskip 0.01in}c}
Image&Flow&SAN&+cond.&DivA&Groundtruth\\
\includegraphics[width=\fWidD]{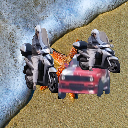}&
\includegraphics[width=\fWidD]{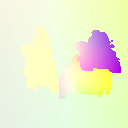}&
\includegraphics[width=\fWidD]{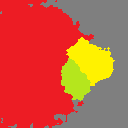}&
\includegraphics[width=\fWidD]{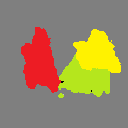}&
\includegraphics[width=\fWidD]{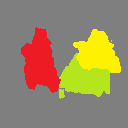}&
\includegraphics[width=\fWidD]{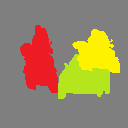}\\
\includegraphics[width=\fWidD]{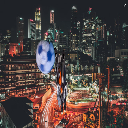}&
\includegraphics[width=\fWidD]{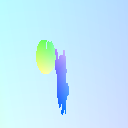}&
\includegraphics[width=\fWidD]{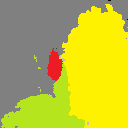}&
\includegraphics[width=\fWidD]{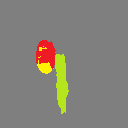}&
\includegraphics[width=\fWidD]{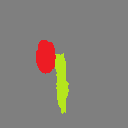}&
\includegraphics[width=\fWidD]{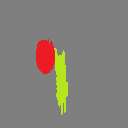}\\
\includegraphics[width=\fWidD]{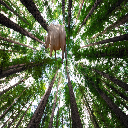}&
\includegraphics[width=\fWidD]{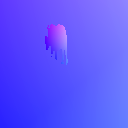}&
\includegraphics[width=\fWidD]{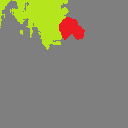}&
\includegraphics[width=\fWidD]{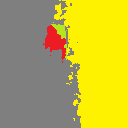}&
\includegraphics[width=\fWidD]{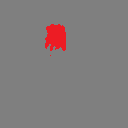}&
\includegraphics[width=\fWidD]{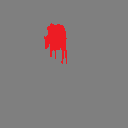}
\end{tabular}
}
\caption{\sl\small {\bf Representative results on diagnostic data.} Vanilla slot attention model can roughly capture moving objects but fails to output accurate segmentation. With conditional decoder, masks align better to object boundaries, but multiple slots may map to the same object. Incorporating adversarial training promotes greater information separation between slots.}
\label{fig:diagnostic}
\end{figure}

\textbf{Empty Slots. }
In cases where the number of slots is more than the number of independent motion patterns in the input, certain slots may remain unallocated to any region in the image domain. These slots do not carry any information about the input flow, and thus the flow reconstructed from these slots are random patterns and the corresponding masks are close to zero. Future work will focus on dynamically allocating the number of slots so that they adapt to the input data, especially in videos where objects appear and disappear as a result of occlusions. 

\textbf{Recursive Implementation. }
We provide a recursive implementation of the algorithm in which slots within the current frame are initialized with corresponding slots from prior frames, and updated by 1 iteration instead of 3. This approach makes the model less vulnerable to label-flipping provided the motion pattern is consistent across frames. However, when the motion undergoes drastic changes, for example when an object stops moving suddenly and then resumes after a hiatus, the model fails to associate the object to the same slot. In order to address this issue, future work on dynamic management of slots will also address such long-term consistency issues. 

\textbf{Static scene.}
In this work, our goal is to discover objects based on motion signals. One question arises: can the proposed method effectively find objects in static scenes? To address this, we present an example in Fig.~\ref{fig:static}. Even in a static scene, changes in viewpoint can lead to a piecewise smooth optical flow, which our method, DivA, can segment. According to Gibson's concept of ``detached objects'' as ``layouts of surfaces surrounded by the medium,'' camera motion can create occlusion boundaries in any scene that is not globally convex. As a result, one moving agent can observe piecewise smooth optical flow with occluding boundaries, induced by either viewer or object motion. In such scenarios, DivA demonstrates its ability to identify objects when embedded in a moving agent, such as a robot, even in a static scene.

\def\figd{figures/rebuttal}
\def\fWidD{0.18\textwidth}
\begin{figure}[t]
\centering
\hspace*{-3mm}
{\footnotesize
\begin{tabular}[0mm]{c@{\hskip 0.01in}c@{\hskip 0.01in}c}
Image Sequence&Optical Flow&Segmentation\\
\includegraphics[width=\fWidD]{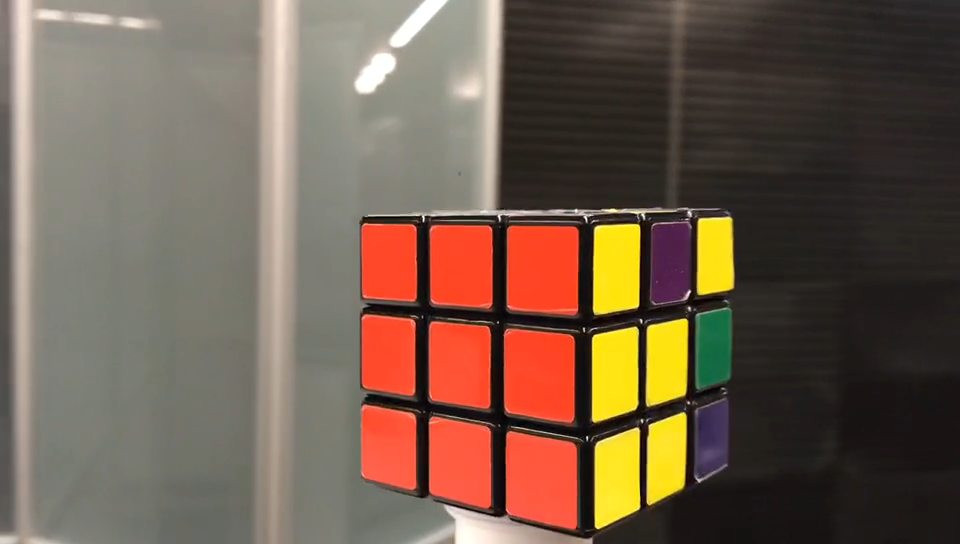}&
\includegraphics[width=\fWidD]{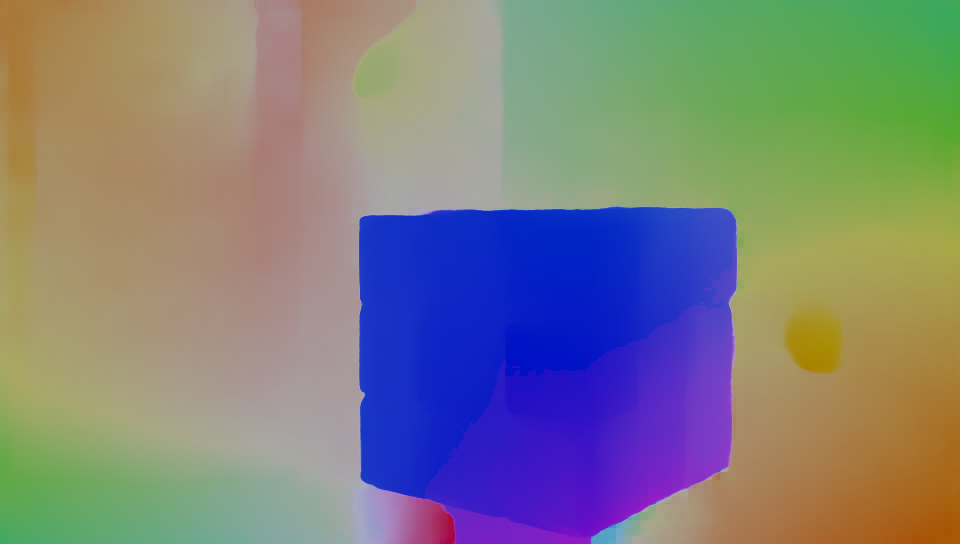}&
\includegraphics[width=\fWidD]{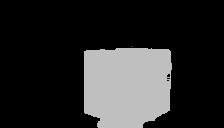}
\end{tabular}   
}
\caption{\sl\small {\bf Segmenting a static scene.} In a static scene, changes in viewpoint can lead to a piecewise smooth optical flow, which can be segmented by DivA.}
\label{fig:static}
\end{figure}

\textbf{Negative Results. }
The main paper shows that adversarial training in DivA functions as an implicit entropy regularization. However, our attempt to integrate the  entropy loss directly into the overall loss 
\begin{equation}
\ell_{\text{entropy}}(u, I) = \sum_{i=1}^n \big\|m_i \log(m_i)\big\|
\end{equation}
is actually detrimental of overall performance. 
We speculate that the entropy loss solely stimulates binary segmentation output and does not mandate any form of information separation. Incorporating this loss may keep the model from capturing the motion pattern of the complete object. Additional work will be needed to analyze the regularizing role of the adversarial loss and compare it to explicit regularization forms that are believed to foster information separation of ``disentanglement.''

\end{document}